\documentclass[conference]{IEEEtran}

\IEEEoverridecommandlockouts  

\usepackage{amsfonts,amssymb,amsmath,mathtools,multicol}
\usepackage{physics}

\usepackage{float}
\usepackage{graphicx}
\graphicspath{{images/}}

\usepackage{cite}
\usepackage{hyperref}
\usepackage{subcaption}
\usepackage[dvipsnames]{xcolor}

\usepackage{algorithm}
\usepackage[indLines=true, 
		     noEnd=true, 
		     beginLComment=//~, 
		     endLComment=~, 
		     beginComment=//~]{algpseudocodex}
\renewcommand{\algorithmicrequire}{\textbf{Input:}}
\renewcommand{\algorithmicensure}{\textbf{Output:}}

\usepackage[inline]{enumitem}
\usepackage{cancel}

\usepackage{tabularx}   
\usepackage{booktabs}   
\usepackage{multirow}

\newcommand{\V}[1]{{\boldsymbol{\mathbf{#1}}}}

\newcommand{\R}{\mathbb{R}}

\title{Path Planning in Complex Environments with Superquadrics and Voronoi-Based Orientation}
\author{Lin Yang$^{1\dagger}$, Ganesh Iyer$^{2\dagger}$, Baichuan Lou$^{1}$, Sri Harsha Turlapati$^{1}$, Chen Lv$^{1}$, Domenico Campolo$^{1*}$
\thanks{$^{1}$ Authors are with the School of Mechanical and Aerospace Engineering, Nanyang Technological University (NTU), Singapore.}
\thanks{$^{2}$ Authors are with the School of Mechanical Engineering, Indian Institute of Technology, Bombay, India.}
\thanks{$^{\dagger}$ These authors contributed equally to this work}
\thanks{$^*$ Corresponding author: {\tt d.campolo@ntu.edu.sg}}
\thanks{This research is supported by the National Research Foundation, Singapore, under the NRF Medium Sized Centre scheme (CARTIN).}
}
\date{}

\begin{document}
\maketitle

\begin{abstract} 
Path planning in narrow passages is a challenging problem in various applications. Traditional planning algorithms often struggle in complex environments such as mazes and traps, where navigating through narrow entrances requires precise orientation control. Additionally, perception errors can lead to collisions if the planned path is too close to obstacles. Maintaining a safe distance from obstacles can enhance navigation safety.
In this work, we present a novel approach that combines superquadrics (SQ) representation and Voronoi diagrams to solve the narrow passage problem. The key contributions are: $i)$ an efficient and generalizable path planning framework, integrating SQ with Voronoi diagrams to provide a computationally efficient and differentiable mathematical formulation, enabling flexible shape fitting for various robots and obstacles. $ii)$ improved passage feasibility, achieved by expanding SQ’s minor axis to eliminate impassable narrow gaps while maintaining feasible paths along the major axis, which aligns with the Voronoi hyperplane and follows maximum clearance paths. $iii)$ enhanced robustness in complex environments, by integrating multiple connected obstacles into the same Voronoi region, handling trap scenarios.
We validate our framework through a 2D object retrieval task and 3D drone simulation, demonstrating that our approach outperforms classical planners and a cutting-edge drone planner in various key aspects. The video can be found at \url{https://youtu.be/kbvQRxRACvQ}
\end{abstract}


\begin{IEEEkeywords}
Voronoi diagrams, Superquadrics, Motion planning, Narrow passage.
\end{IEEEkeywords}

\section{Introduction} 

\begin{figure}[!h]
    \centering
    \begin{subfigure}[b]{0.2\textwidth}
        \centering
        \includegraphics[width=\textwidth]{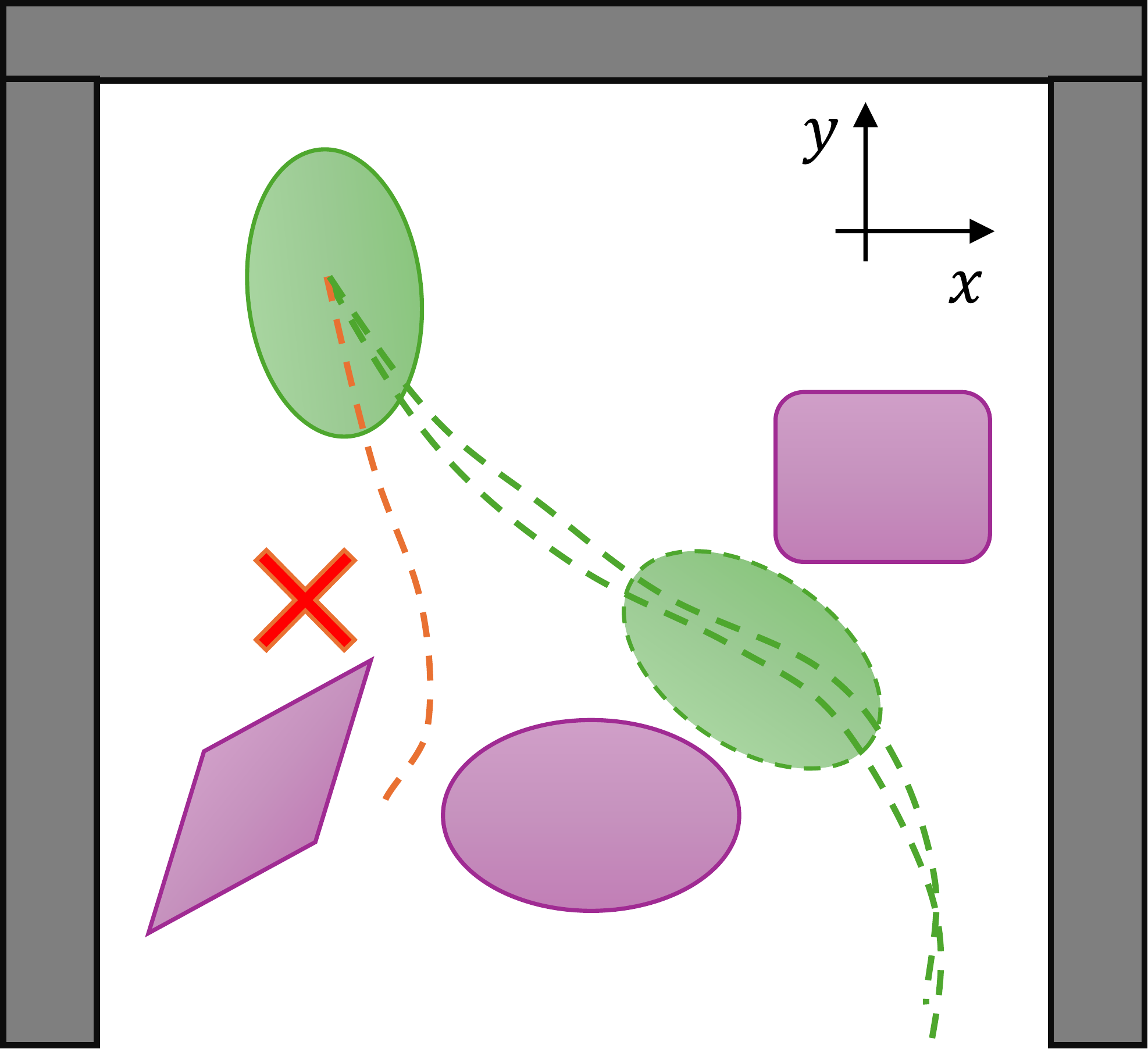}
        \caption[Network2]%
        {{\small Planning in deterministic case: identifying passable and impassable routes}}    
        \label{fig1:a}
    \end{subfigure}
    \hfill
    \begin{subfigure}[b]{0.2\textwidth}  
        \centering 
        \includegraphics[width=\textwidth]{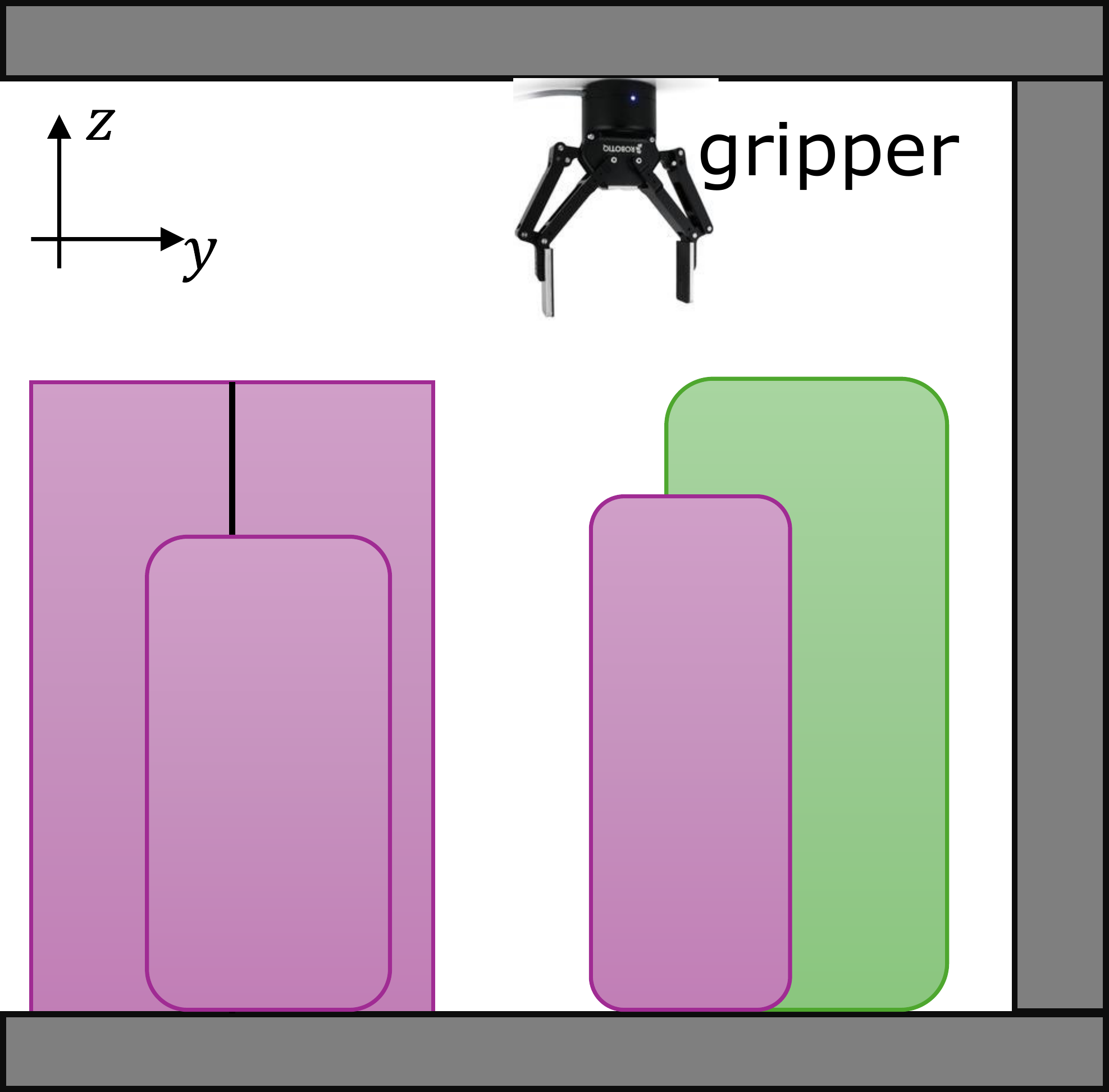}
        \caption[]%
        {{\small Object retrieval in cluttered spaces: cross-sectional view}}    
        \label{fig1:b}
    \end{subfigure}
    \vskip\baselineskip
    \begin{subfigure}[b]{0.2\textwidth}   
        \centering 
        \includegraphics[width=\textwidth]{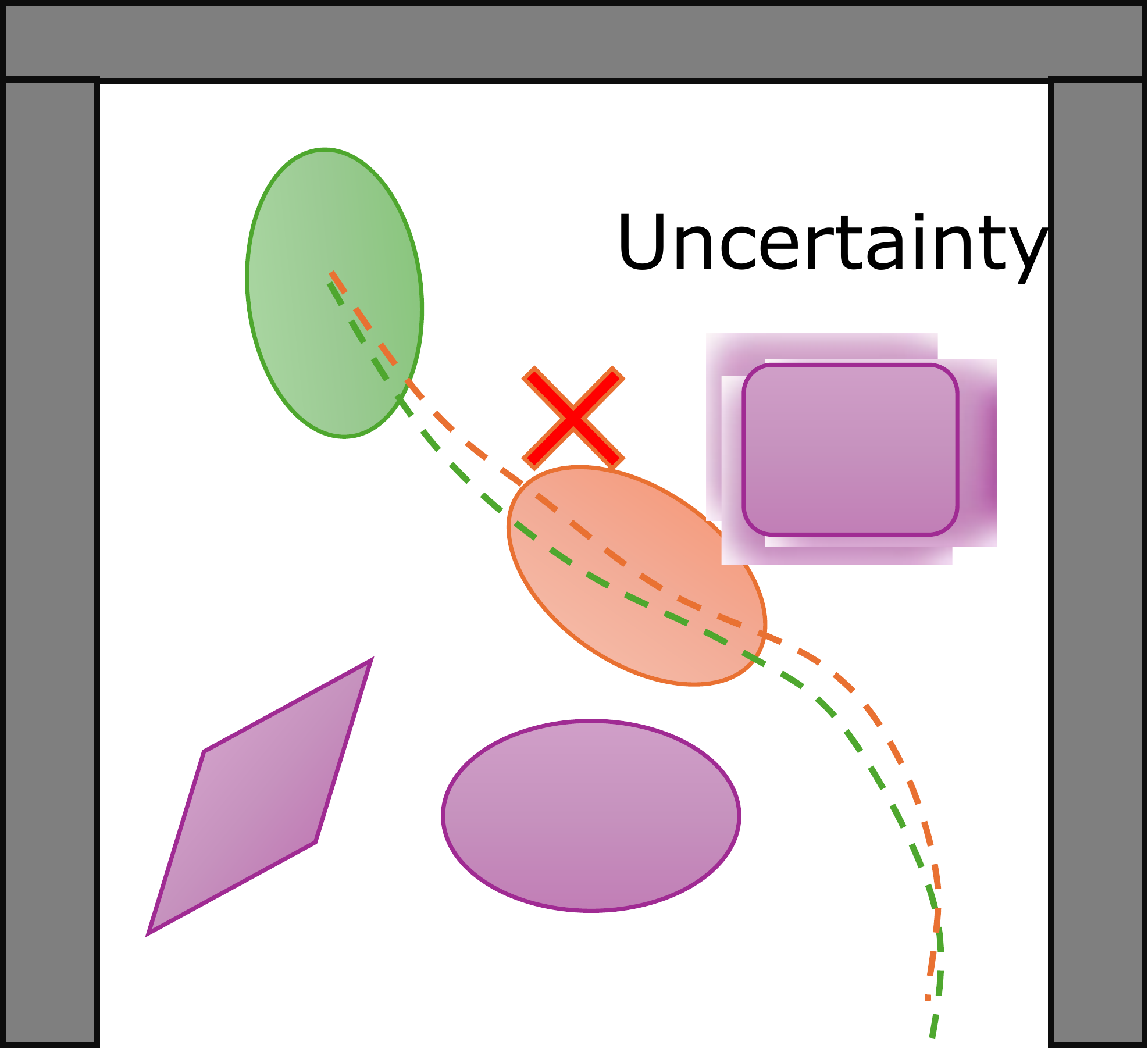}
        \caption[]%
        {{\small Uncertainty in real-world: potential collisions due to perception errors}} 
        \label{fig1:c}
    \end{subfigure}
    \hfill
    \begin{subfigure}[b]{0.2\textwidth}   
        \centering 
        \includegraphics[width=\textwidth]{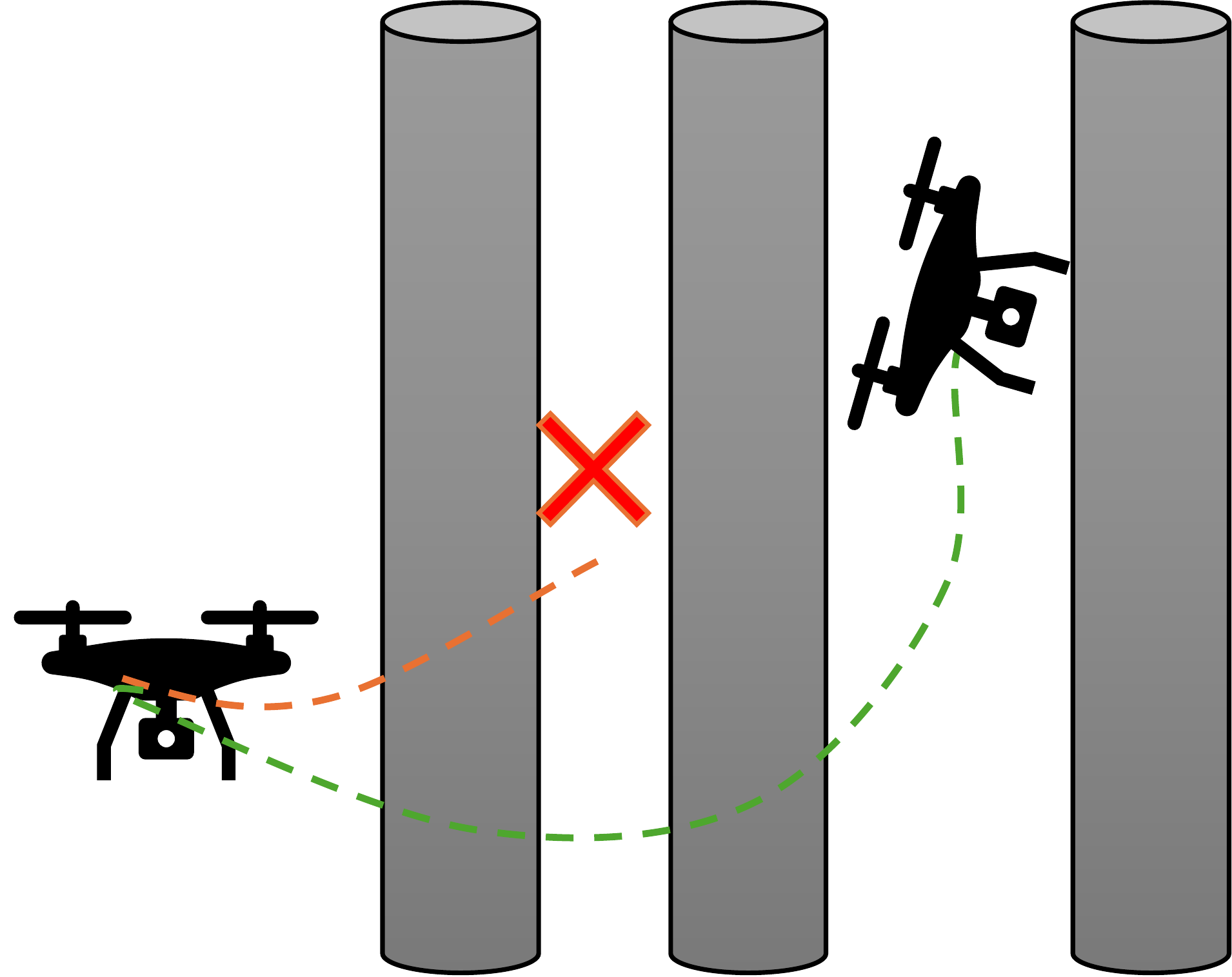}
        \caption[]%
        {{\small 3D Navigation through narrow passages: a drone pass through narrow pillars}}    
        \label{fig1:d}
    \end{subfigure}
    \caption[]
    {\small Navigating narrow passages in 2D and 3D: path feasibility, orientation decision, and safety problem due to uncertainty} 
    \label{fig1}
\end{figure}

Path planning in complex environment such as narrow passages and traps poses challenges across various applications, including robotic manipulation \cite{yang2023planning}, mobile robotics \cite{chen2023set}, and drone navigation \cite{gao2020teach}. The task involves guiding a robot to navigate through a narrow passage without colliding with obstacles.
The challenge arises from that not all robots or obstacles can be approximated as spheres or grid, though this is a common simplification in mobile robots. This approximation allows for easier handling of collision detection and path planning algorithms, as it reduces the complexity of the vehicle's geometry. However, in many cases, robots must find a path through the passage while maintaining a correct orientation.
For example, as depicted in Fig. \ref{fig1:b}, a manipulator attempts to retrieve an object from a crowded shelf. Due to height constraints, the robot is restricted to navigating in a 2D plane, as shown in Fig. \ref{fig1:a}. Some passages may be too narrow for the robot to traverse, while others are wide enough to accommodate the object's shorter axis length.
However, due to perception uncertainties or control inaccuracies, some of these trajectories may result in collisions with obstacles if they are too close to them (Fig.\ref{fig1:c}). Similarly, in drone navigation, certain passages may require the drone to tilt at specific angles to pass through, while in other cases, the robot must recognize that certain passages are impassable, as illustrated in Fig.\ref{fig1:d}.

Many traditional path planning algorithms are popular and widely applied due to their favorable performance. However, some fall short when dealing with complex environments like mazes, traps, and narrow passages.
For example, while sampling-based planning methods perform well in high-dimensional tasks, they struggle with efficiently sampling and growing a path through narrow passages \cite{jimenez2024visualizing}. 
Similarly, artificial potential field methods are widely used for obstacle avoidance \cite{mac2016heuristic}, but they are prone to getting stuck in local optima.  
One effective approach for navigating narrow passages and mazes is the Voronoi diagram \cite{lei2022intention}. This method provides the safest solution for the planner by ensuring maximum clearance, which means the robot remains as far as possible from all obstacles.
Additionally, some algorithms have already achieved near real-time efficiency in mobile robot path planning \cite{arslan2019sensor} with the help of Voronoi diagram. However, these approaches lack pose control in narrow passage scenarios because circular mobile robots typically do not need orientation control in such environments.


The choice to approximate objects is crucial. A model that is too simplified (e.g., a sphere) may prevent the robot from passing through certain narrow passages.
For instance, if the green ellipse in Fig. \ref{fig1:a} is approximated as a sphere, the robot would lose the opportunity to retrieve the object. On the other hand, using an accurate model can result in computational overhead during collision detection.
Orthey et al. \cite{orthey2021section} successfully tackled narrow passage problems by employing an efficient yet complex approach. Their method, while powerful, requires numerous patterns and a detailed CAD model of the object for collision detection.

In contrast, we prioritize a balance between simplification and efficiency by representing objects using superquadrics (SQ) \cite{jaklic2000segmentation}, which provide both differentiability and a convenient parametrization.
SQ can describe a wide range of geometric shapes. For example, Ruan et al. \cite{ruan2022efficient} address narrow passage problems using SQ to represent both robots and obstacles, employing Minkowski sums to avoid collisions across various robot orientations. Their approach achieved effective planning for self-collision avoidance.
Zhao et al. \cite{zhao2022solving} simplify the Minkowski sum via utilizing the formulation of SQ.
For now, our task is simpler than \cite{ruan2022efficient} and thus, we do not require extensive computation with Minkowski sums. Additionally, we avoid over-simplification, as many researchers use Minkowski sums to grow obstacle sizes by sweeping the maximum size of the robot, ensuring the remaining map is navigable with any robot pose.

In this work, we propose an efficient approach for addressing the narrow passage problem in both 2D and 3D scenarios by integrating SQ representation with Voronoi diagrams. Our method offers three key advantages:
\begin{enumerate}[label=\roman*)]
  \item \textbf{An efficient path planning framework:} 
SQs can model various geometries, ensuring adaptability to various shapes while maintaining differentiability.
Parameterizing SQs in polar coordinates enables efficient hyperplane computation and Voronoi diagram generation.
  \item \textbf{Improved passage feasibility and orientation guidance:}
  Expanding the SQ’s minor axis eliminates impassable gaps while preserving feasible paths.
Aligning with Voronoi hyperplanes ensures maximum clearance paths and naturally guides the robot’s orientation.
  \item \textbf{Enhanced robustness in complex environments:} Segmenting connected obstacles into the same Voronoi region prevents the robot from getting trapped. Validated in 2D and 3D scenarios, our method successfully identifies the correct passage and orientation while avoiding traps.
In 3D simulations, it outperforms a cutting-edge drone planner, finding passable paths in a narrow-passage world.
\end{enumerate}

\section{Preliminaries: Superquadrics and Voronoi diagram} \label{SQVD}
\begin{figure*}[t]  
  \begin{subfigure}{0.2\textwidth}
    \includegraphics[width=\linewidth]{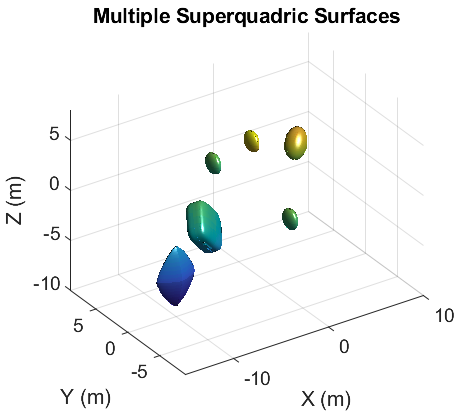}
    \caption[]
    {\small  Original obstacles.}
    \label{SUBFIG:1}
  \end{subfigure}%
  \hfill  
  \begin{subfigure}{0.2\textwidth}
    \includegraphics[width=\linewidth]{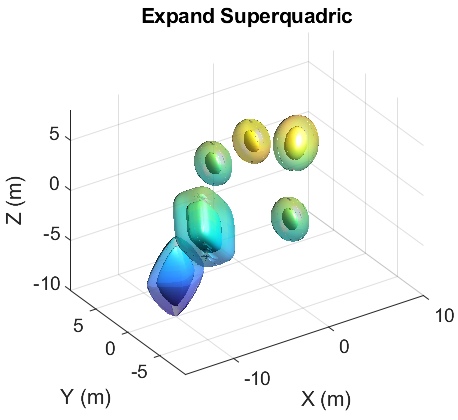}
    \caption[] 
    {\small Grown obstacles.}
    \label{SUBFIG:2}
  \end{subfigure}%
  \hfill  
  \begin{subfigure}{0.2\textwidth}
    \includegraphics[width=\linewidth]{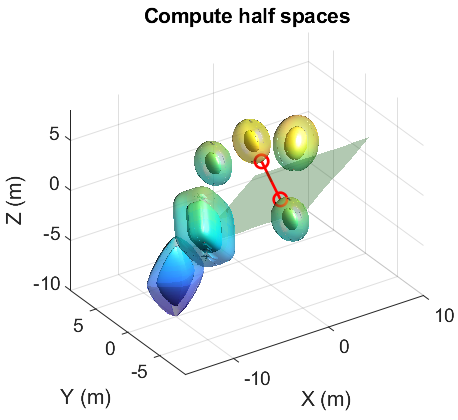}
    \caption[] 
    {\small Hyperplane.}
    \label{SUBFIG:3}
  \end{subfigure}%
    \hfill  
  \begin{subfigure}{0.2\textwidth}
    \includegraphics[width=\linewidth]{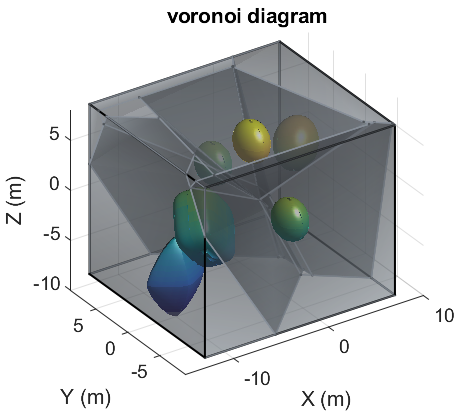}
    \caption[] 
    {\small Create Voronoi diagram.}
    \label{SUBFIG:4}
  \end{subfigure}%
    \hfill  
  \begin{subfigure}{0.2\textwidth}
    \includegraphics[width=\linewidth]{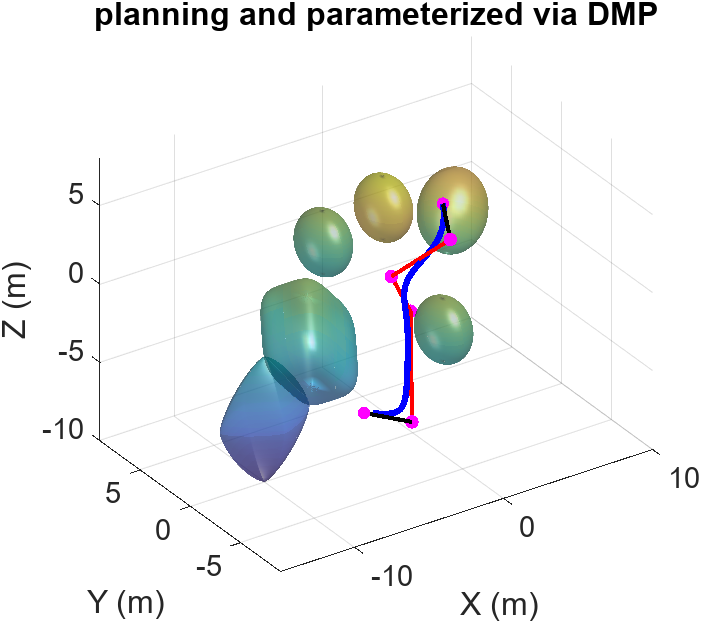}
    \caption[] 
    {\small planning via DMP.}
    \label{SUBFIG:5}
  \end{subfigure}%
\caption{An entire process of our algorithm: a): approximate the environment by SQs. b): Grown obstacles with shortest axis length of robot. c): Hyperplane between two SQs and closest point. d): Create Voronoi diagram with hyperplane. e): Utilize graph planning to find edges, then parameterized by DMP.} 
\end{figure*}

In this section, we present the mathematical definitions of Superquadrics (SQ) (known as Superellipses in 2D) and the Voronoi diagram. Superquadrics form a family of three-dimensional shapes that generalize basic quadric surfaces, such as ellipsoids, elliptic paraboloids, and hyperbolic paraboloids. These surfaces are generated by the spherical product of two parametric two-dimensional curves \cite{jaklic2000segmentation}.

\subsection{Superellipses and superquadrics}

Superquadrics are widely utilized to represent geometry of objects in real world \cite{liu2022robust}. The 2D analogue of superquadrics are superellipses,

\subsubsection{Superellipses}

Superellipses is implicitly defined via equation:
\begin{align}
    \left(\frac{x}{a_1}\right)^{\frac{2}{\epsilon}} + \left(\frac{y}{a_2}\right)^{\frac{2}{\epsilon}} = 1
\end{align}
Here, $\epsilon$ represents the exponent of the superellipse, which controls the shape of the superellipse. $a_1, a_2$ control the size. 
Meanwhile, define a inside-outside function $F(x,y)$ as 
\begin{align}
    F(x,y) &= \left(\frac{x}{a_1}\right)^{\frac{2}{\epsilon}} + \left(\frac{y}{a_2}\right)^{\frac{2}{\epsilon}} - 1
    \label{eq:io}
\end{align}
The function $F(x,y)$ possesses a very useful property. For a point $(x_0,y_0)$, if $F(x_0,y_0) > 0$, then it is outside the SQ, if $F(x_0,y_0) = 0$ then it is on the surface, and if $F(x_0,y_0) < 0$, then it is inside. For this reason, the function F is also referred to as the \textbf{inside-outside} function.

Meanwhile, analogous to a circle, superellipse can be parameterized as \cite{jaklic2000segmentation},

\begin{subequations}
\begin{align}
    \mathbf{r}(\omega) & = \begin{bmatrix}
    a_1 \cos^\epsilon \omega \\
    a_2 \sin^\epsilon \omega
    \end{bmatrix}, \; \omega \in [-\pi,\pi]
\end{align}
\end{subequations}

The exponentiation with power $\epsilon$ here is the signed power function, i.e 
\begin{align}
   \cos^\epsilon \omega = \operatorname{sign}(\cos\omega) \cdot \left| \cos\omega \right|^{\epsilon}
\end{align}

\subsubsection{Superquadrics}
In 3D case, the implicit function of superquadrics is defined as,
\begin{align}
    \left(\left(\frac{x}{a_1}\right)^{2/\epsilon_2} + \left(\frac{y}{a_2}\right)^{2/\epsilon_2} \right)^{\epsilon_2/\epsilon_1} + \left(\frac{z}{a_3}\right)^{2/\epsilon_1} &= 1
    \label{eq:F3D}
\end{align}
So we have a similar inside-outside function $F(x,y,z)$ (analgoy to Eq. \ref{eq:io} for 3D case)
\begin{align}
    F(x,y,z) &= \left(\left(\frac{x}{a_1}\right)^{2/\epsilon_2} + \left(\frac{y}{a_2}\right)^{2/\epsilon_2} \right)^{\epsilon_2/\epsilon_1} + \left(\frac{z}{a_3}\right)^{2/\epsilon_1} -1
    \label{eq:in_out}
\end{align}
$F(x,y,z)$ is still \textbf{inside-outside} function in 3D case.
Meanwhile, Eq. \ref{eq:F3D} can be parameterized by 
\begin{subequations}
\begin{align}
    \mathbf{r}(\eta,\omega) & = \begin{bmatrix}
    a_1cos^{\epsilon_1}\eta ~cos^{\epsilon_2}\omega \\ 
    a_2cos^{\epsilon_1}\eta~sin^{\epsilon_2}\omega\\
    a_3sin^{\epsilon_1}\eta \end{bmatrix}, \; \eta \in [-\pi/2,\pi/2],~\omega \in [-\pi,\pi]
\end{align}
\end{subequations}
Here, $(a_1,a_2,a_3)$ are the size parameters, while $(\epsilon_1,\epsilon_2)$ are the shape parameters. Thus, in local coordinates of the superquadric, 5 parameters are needed to define it completely.


\subsubsection{Superellipses and Superquadrics in general position}

To accommodate any orientation and position, translation and rotation operations must be applied to the superellipses and superquadrics. Transformation matrices are required to represent superellipses and superquadrics in the world frame. Therefore, to fully describe a SQ, the parameter set is defined as 
$SQ: = [\V \epsilon, \V a, \V p, \V \theta]^T$, where $\V T(\V p, \V \theta) = [\V R(\V \theta) \in SO(3), \V p \in \R^3] \in SE(3)$ is the transformation matrix.
\vspace{-3mm}
\subsection{Voronoi Diagrams} 

Fundamental data structures play a pivotal role in computational geometry, enabling efficient algorithms for solving various geometric problems. One such essential structure is the Voronoi diagram, which partitions the Euclidean space $W \in \R^n$ into $m$ half spaces $\nu = \{V_1, V_2, \ldots, V_m\}$, often referred to as cells or regions,

Each cell is defined such that every point within it is closest to the corresponding disjoint convex obstacles $\mathcal{O} = \{O_1, O_2, \ldots, O_m\}$ (i.e., $d(O_i, O_j)>0 \ \forall i \neq j$), based on maximum margin separating hyperplane, to be
\begin{subequations}
\begin{align}
    HP_{i,j} &:= \left\{ q \in W \ \middle|\ \|q - p_i^*\| = \|q - p_j^*\| \right\},    \\
    V_i &:= \left\{ q \in W \ \middle|\ \|q - p_i^*\| \leq \|q - p_j^*\| \right\}, \\
    (p_i^*, p_j^*) &= \underset{p_i \in \mathcal{O}_i, p_j \in \mathcal{O}_j}{\arg \min }  d(p_i, p_j) \quad \forall j \neq i
\end{align}
\label{eq:vo}
\end{subequations}
where $HP_{i,j}$ denotes the hyperplane (or boundary) shared by two Voronoi regions $V_i$ and $V_j$. Traveling along this hyperplane ensures that the path remains equidistant from the obstacles associated with $V_i$ and $V_j$, thus providing a locally maximal clearance from these obstacles \cite{arslan2019sensor}.




\section{ robot navigation in generalized world}\label{method}

Many objects can be approximated by one or several SQs \cite{liu2022robust}. Unlike many approaches that assume all obstacles are disjoint and that the distance between obstacles is greater than the robot's maximum diameter, our SQ-based representation allows for a more generalized and realistic world.
 
\vspace{-1mm}
\subsection{Eliminate impossible passage by expanding}
In robot planning, mobile robots \cite{arslan2019sensor} and drones \cite{WANG2022GCOPTER} are commonly approximated by a sphere, after which collision checks or Minkowski sums are applied to all obstacles.

However, this method restricts the ability of non-spherical robots, such as a truck or a peg-like object, from passing through narrow passages. Therefore, in our approach, the first step is to approximate the robot, object, or drone using a Superquadric, represented as $SQ_r = [\V \epsilon_r, \V a_r, \V p_r, \V \theta_r]^T$, while the other obstacles are represented as $SQ_o = [\V \epsilon_o, \V a_o, \V p_o, \V \theta_o]^T$. For example, approximate some obstacles like Fig. \ref{SUBFIG:1}.

We arrange the components of $a_i$ in ascending order, denoted as $\V a =[a_1, a_2, a_3], a_i<a_{i+1}$
so $a_{r,1}$ is the shortest axis of $SQ_r$. We expand this shortest axis to evaluate whether the robot can navigate through a narrow passage.

While the original concept of Voronoi-based planning is effective for point robots, it cannot be directly applied to robots with extended bodies. The Minkowski sum, a fundamental operation in computational geometry and robotics, is commonly used to handle this situation.
Given two sets \(A\) and \(B\) in a vector space, the Minkowski sum \(A + B\) is defined as the set of all points that can be expressed as the sum of a point from \(A\) and a point from \(B\): \(A+B = \{a+b \ | \ a \in A, b \in B\}\). Instead of applying the Minkowski sum directly, we leverage the properties of Superquadrics to achieve similar results by expanding $\V a_o$ to achieve similar performance of Minkowski sum. This expansion yields an adjusted obstacle, represented $SQ^M_o$, where the size parameters are modified accordingly to reflect the expanded configuration.
\begin{align}
    \V a^M & = \V a +  a_{r,1} \V I 
    \label{eq:expand}
\end{align}
where $\V I$ denotes the identity matrix. Now the robot is reduced to a point, as shown in Fig. \ref{SUBFIG:2}, on which we can apply standard voronoi diagram.

\subsection{Generate Voronoi diagram}

One advantage of using SQ is that they can be parameterized by angular coordinates, making them continuous and differentiable functions. This property allows us to efficiently compute the minimum distance between SQ pairs using classical optimization algorithms (e.g., Newton's method) to identify the closest points between each SQ pair.
As illustrated in Fig. \ref{SUBFIG:3}, two obstacles are represented by two SQs. The red circles indicate the points $(p_i^*, p_j^*)$, where the distance $d(p_i, p_j)$ is minimized (Eq. \ref{eq:vo}), and the green plane represents the hyperplane that separates the two regions.

Meanwhile, we evaluate whether $F_i(p_j^*)\leq 0$. If this condition holds, the two SQs overlap, and we assign them to the same cluster $C_i$. Otherwise, we separate two SQ as two clusters $C_i$ and $C_j$. The logic is implemented in Alg. \ref{alg: cluster}.

\begin{algorithm}[!ht]
\caption{Generate cluster}\label{alg: cluster}
\begin{algorithmic}[1]
\Require Obstacles after mink sum $\V SQ^M_o $, the world space $W$.
\Procedure{Create Cluster}{$\V SQ^M_o, W$}
\State Cluster $C \gets\emptyset$, cluster index $id=1$.
\For{$i=1,\dots,N$}
 \If{$SQ^M_{o,i} \notin C$}
	\State $id \gets id +1$
        \State $C_{id} \gets \emptyset$
\EndIf
\For{$j=i,\dots,N$}
	\State$(p_i^*, p_j^*)\gets \underset{p_i \in SQ^M_{o,i}, p_j \in SQ^M_{o,j}}{\arg \min }  d(p_i, p_j)$,
 
 \If{$F_i(p_j^*)\leq 0$}
	\State $C_i \gets C_i \cup SQ^M_{o,i}$
\EndIf
\EndFor
\State $C \gets C \cup C_i$
\EndFor
\State\Return $C$
\EndProcedure
\end{algorithmic}
\end{algorithm}

After separating the clusters based on the overlapping of SQs, we proceed to compute the elements in Eq. \ref{eq:vo} to create the diagram.
For any the hyperplane $HP_{i,j}$ between SQ $i$ and SQ $j$ can be expressed at the center of the $\frac{p^*_i+p^*_j}{2}$. Meanwhile, define the normalize function $\hat{n}(x)$ as
\begin{align}
    \hat{n}(\V x) = \frac{\V x}{\norm{\V x}}
\end{align}
Thus, the normal vector of $HP_{i,j}$ equals,
\begin{align}
    \V n_{i,j} = \hat{n}(\V p^*_i-\V p^*_j)
\end{align}
Following the definition of hyperplane, we get 
\begin{align}
    HP_{i,j} = \{x | \V n_{i,j}^T x \leq \V n_{i,j}^T (\frac{\V p^*_i+\V p^*_j}{2})\}
    \label{eq:HP}
\end{align}
A property of voronoi diagrams is that, for each object $SQ_o$, the intersection of all the spaces created by the bisector hyperplanes that contain the object forms the polyhedral cell corresponding to that object. The collection of all such polyhedral cells constitutes the complete Voronoi diagram.
Initially, We define a bounding box (world space $W$) to appropriately represent the workspace. Consequently, the region $V_i$ corresponding to each cluster is represented as the intersection of all the hyperplanes that define its polyhedral cell within this bounding box.
\begin{align}
   V_i =  \bigcup_{j=0}^{m} W \cap HP_{i,j}.
   \label{eq:V_i}
\end{align}
The result of Eq. \ref{eq:V_i} is shown as Fig.\ref{SUBFIG:4}. We observe most obstacles belong to one region, while there are two overlap obstacles belongs to same region.
The steps to generate a Voronoi diagram with cluster is shown in Alg. \ref{alg: Voronoi}. 
\begin{algorithm}[!ht]
\caption{Generate Voronoi diagram}\label{alg: Voronoi}
\begin{algorithmic}[1]
\Require Cluster $C = {[C_1, \dots, C_K]}$, the world space $W$.
\Procedure{Create Voronoi}{$C, W$}
\For{$i=1,\dots, K$}
\State $C_i = C(i)$
\For{$j=i,\dots,K$}
\State $C_j = C(j)$
 \If{$i \neq j$}
	\State$(p_i^*, p_j^*)\gets \underset{\V p_i, \V p_j}{\arg \min } \; d(\V p_i, \V p_j)$, where $\V p_i \in SQ^M_{o,i} \in C_i, \; \V p_j\in SQ^M_{o,j} \in C_j$
\EndIf
	\State $HP_{i,j}\gets$ Eq. \ref{eq:HP}.
    \State $V_i =  \bigcup_{j=0}^{m} W \cap HP_{i,j}$ Eq. \ref{eq:V_i}.
\EndFor
\EndFor
\State\Return $V$
\EndProcedure
\end{algorithmic}
\end{algorithm}

The diagrams illustrating the performance of Alg. \ref{alg: cluster} and Alg. \ref{alg: Voronoi} are shown below

\begin{figure}[!ht]
\centering
\includegraphics[width=0.45\textwidth]{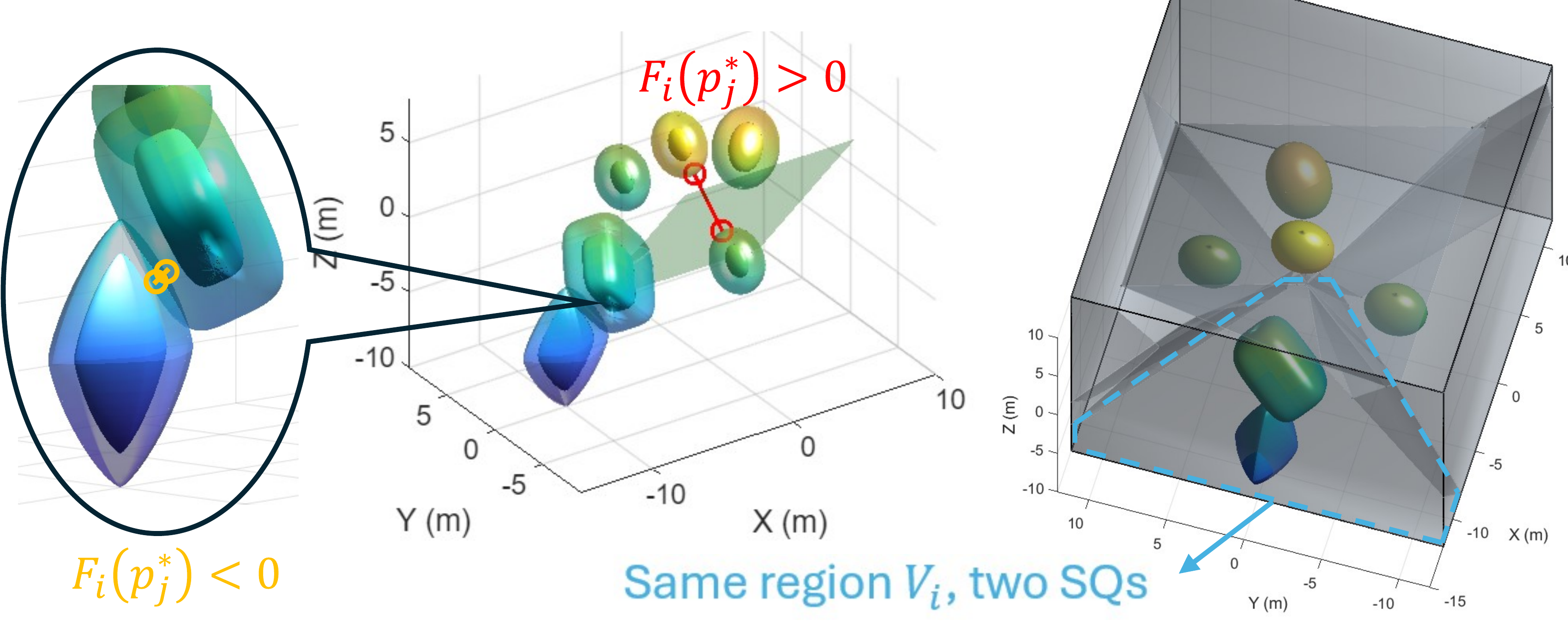}
\caption{Clusters are partitioned using Alg. \ref{alg: cluster}, and adjacent SQs are grouped into the same region as Alg. \ref{alg: Voronoi}.}
\label{fig: alg12}
\end{figure}

\vspace{-3mm}
\subsection{Generate graph via Voronoi diagram}
We create a graph $\mathcal{G}=(\mathcal V, \mathcal E)$,
where each $V_i$ corresponds to a polygon with multiple vertices and edges. Each vertex of the polygon is represented as a node $v_i$ in a graph $\mathcal V$, where nodes are connected by edge on polygon $\mathcal E$. The weight assigned to each edge corresponds to the Euclidean distance between the connected nodes.

We notice that there are certain regions (holes) in the diagram that do not belong to any cell $V_i$. This occurs due to the linear approximation used in constructing the diagram, but it has few impact since our objective is to identify safe paths along the roadmap. 
Thus, we connect nodes where the distance between them is smaller than a predefined threshold $h$, which is determined based on the overall size of the workspace. In other words, this is used to connect the nodes near the hole to maintain a practical and navigable graph structure.


\begin{align}
    \mathcal E & = \mathcal E \cup \{v_i,v_j\}, \; \forall v_i, v_j \in V, \norm{v_i-v_j} \leq h
\end{align}

Given any initial pose and target pose, we first project these poses onto their respective closest edges $\mathcal E$. Once projected, we determine the shortest path through the graph by classical graph planning algorithm (e.g., Dijkstra) to find solution path via nodes $\mathcal V_{sol} = [v_1, v_2, \dots, v_i]^T$.

\subsection{Planning with orientation}

In subsection A, we determined the longest axis of the robot $SQ_r$. To control the robot's orientation while navigating through narrow passages, the robot should move along the direction of its longest axis. This direction is computed based on the nodes in the path solution, $\mathcal V_{sol}$, which provides the optimal orientation for each step of the path.

\subsubsection{2D}
In the 2D case, we plan the robot's motion using $\V T(\V p, \V \theta) \in SE(2)$, where $\V \theta \in \mathfrak{so}(2)$. For each edge in the path, the direction is determined by the orientation of the robot's longest axis relative to the edge,
\begin{subequations}
    \begin{align}
    \Delta v &:= [\Delta v_x, \Delta v_y]^T = v_{i+1} - v_i \in \R^2 \\
    \V \theta & = \arctan (\Delta v_y, \Delta v_x)
    \end{align}
    \label{eq:ori_2D}
\end{subequations}
\vspace{-3mm}
\subsubsection{3D}
In 3D case, we are planning for $\V T(\V p, \V \theta) \in SE(3)$, where $\V \theta \in \mathfrak{so}(3)$. 
Since we are navigating through nodes located on the surfaces of polyhedra, these surfaces provide the maximum clearance orientation for the robot.
To ensure efficient navigation, the robot's ($SQ_r$) longest axis is aligned forward along the direction of travel, while the shortest axis is oriented perpendicular (normal) to the halfplane $HP_{i,j}$. This configuration maximizes the robot's ability to fit through narrow passages and maintain optimal clearance. The rotation vector required to adjust the robot's orientation is computed based on the vertices of the surfaces, ensuring that the robot aligns itself as it moves from node to node on the surface.
\begin{align}
\V r_1 &= \hat{n}(v_{i+1} - v_i) \nonumber \\
\V r_3 &= \V n_{i,j} , \; v_{i+1}, v_i \in HP_{i,j}  \nonumber \\
\V r_2 &= \V r_3 \times \V r_1 \nonumber \\
\V R &= [\V r_1, \V r_2, \V r_3]
\label{eq:ori_3D}
\end{align}
After determining the rotation matrix at each edge, we compute the corresponding Lie algebra $\V \theta \in \mathfrak{so}(3)$. This is done by applying the matrix logarithm $\log$ to the rotation matrix, which gives a skew-symmetric matrix, then $\vee$ represents extract skew-symmetric matrix to vector.
\begin{align}
\V \theta  &= \log_{\vee} \V R
\end{align}
\subsection{Smooth Interpolation Using DMP}
Now that we have the orientation at each edge and the nodes as via points, we need a smooth trajectory for the robot to follow. First, we interpolate all the nodes using polynomials to obtain the trajectory $\V p(t)$. However, this trajectory may still appear zigzagged, and it is not necessary to pass through every node, especially in areas with many redundant nodes (e.g., holes in the diagram). To address this, we employ Dynamical Movement Primitives (DMP) to learn a smooth version of this trajectory. DMP is a technique that parameterizes and smooths trajectories \cite{ijspeert2013dynamical}. Classical DMP can be used to map a finite dimensional set of parameters $\V W \in\R^{K\times P}$ where $P$ is the number of Radial Basis Functions (RBFs) per degree-of-freedom (DOF) into smooth and differentiable functions.
\begin{align}
\textsc{DMP}&: (\V W, \V u_{start}, \V u_{goal}, T)\mapsto \V u_{\V W}(t),\quad t\in[0,T].
\end{align}
where $\V u_{\V W}(t)$ represents the desired pose.
We first augment position and lie algebra, so we have dataset $\V D = [\V p(t), \V \theta(t)], t\in[0,T]$, then we apply locally weighted regression (LWR) \cite{ijspeert2013dynamical} to learn parameter $\V W $ from dataset $\V D$, which is used to generate the new trajectory.

\begin{align}
\V W  &= \textsc{LWR} (\V D)
\end{align}

The resulting trajectory is shown in Fig. \ref{SUBFIG:5}, where the blue line represents the original edges of the solution path, and the pink trajectory generated by DMP shows the smoothed path.

\section{Experiment: Motion Planning in Cluttered 2D Environments} \label{exp2D}

Retrieving a target object from clutter is a common task in daily life (e.g., from shelves). However, due to height restrictions, retrieving objects from the top is not feasible, making it challenging to find a collision-free path for retrieval \cite{nam2020fast}. We designed a toy experiment to simulate a similar scenario and test the performance of our algorithm in a 2D environment.

\subsection{Experiment design and results}
\begin{figure*}[t]
\centering
\includegraphics[width=0.7\textwidth]{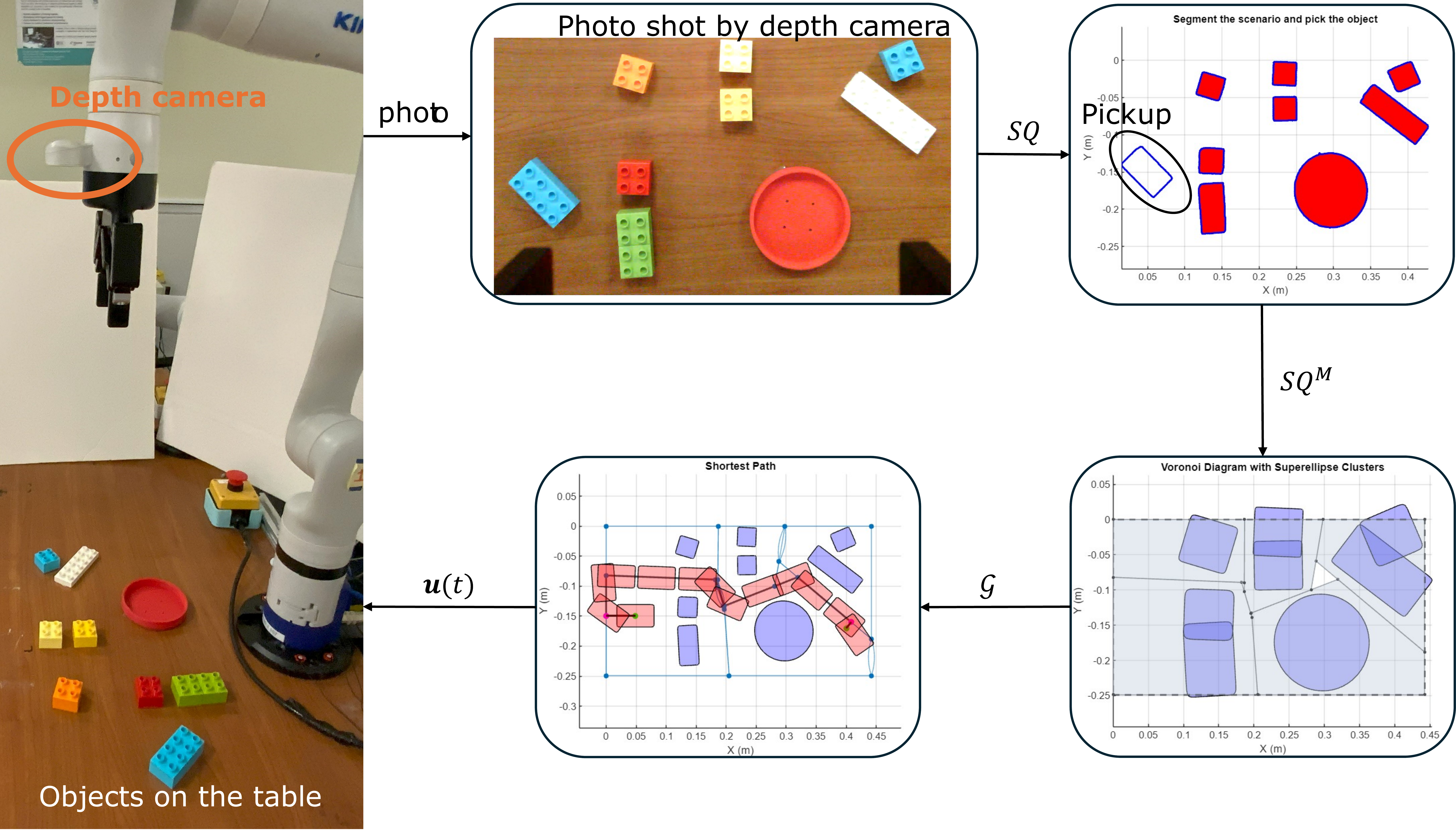}
\caption{Process for executing the narrow-passage experiment. Step 1: The Kinova robot scans a table crowded with objects. Step 2: Objects are segmented, and the target object is selected as $SQ_r$. Step 3: The remaining obstacles $SQ_o$ are expanded, where the impossible passage are eliminated. Step 4: Trajectory planning is performed with the desired orientation.}
\label{fig:exp_2D}
\end{figure*}
The process of this experiment is illustrated in Fig. \ref{fig:exp_2D}. We set up a table with several objects, some of which are placed to create narrow passages that too difficult to pass through. A Kinova robot equipped with a depth camera is positioned on the table, and the camera is calibrated. The robot raises to its highest position while keeping its gaze downwards, and images are captured.

We segment the objects using the SAM toolbox in MATLAB package \cite{kirillov2023segment}, from which we obtain point clouds representing the boundaries of all obstacles. Superellipse fitting is then applied to these point clouds to obtain the obstacle representation \cite{liu2022robust}.
Next, we applied our framework with the help of bensolve toolbox \cite{ciripoi2019calculus}, and one result is shown at Fig. \ref{fig:exp_2D}. It is called narrow passage scenario since some passages between obstacles are impossible to pass through.
We manually select an object $SQ_r$ for retrieval. The robot moves to $\V p_r, \V \theta_r$ to grasp the object. Subsequently in the right bottom figure in Fig. \ref{fig:exp_2D}, the size of other obstacles are expanded by $a_{r,1}$, which eliminates impossible passages.
After expansion, objects close to each other overlap, indicating that they belong to the same Voronoi region $V_i$ (Eq. \ref{eq:V_i}). 
Planning with orientation ensures that the long axis of the object aligns with the edge in Voronoi diagram (Eq. \ref{eq:ori_2D}). The final trajectory is sent to the robot, which executes the trajectory, allowing the object to safely pass through the cluttered environment.

\subsection{Benchmark for 2D scenario} \label{sec:2D}

\begin{figure}[!h]  
  \begin{subfigure}{0.24\textwidth}
    \includegraphics[width=\linewidth]{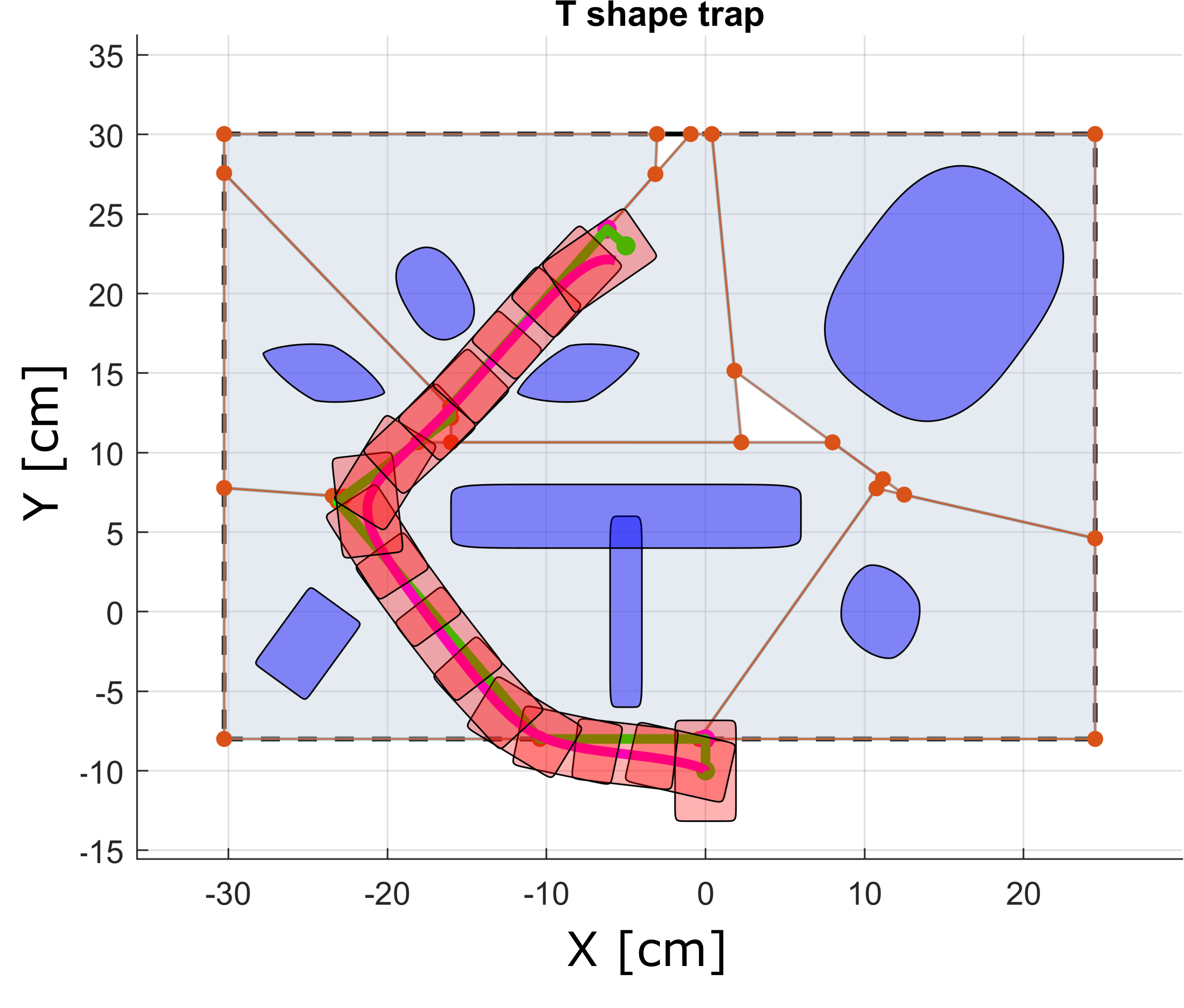}
    \caption[]
    {\small U block scenario.}
  \end{subfigure}%
  \hfill  
  \begin{subfigure}{0.24\textwidth}
    \includegraphics[width=\linewidth]{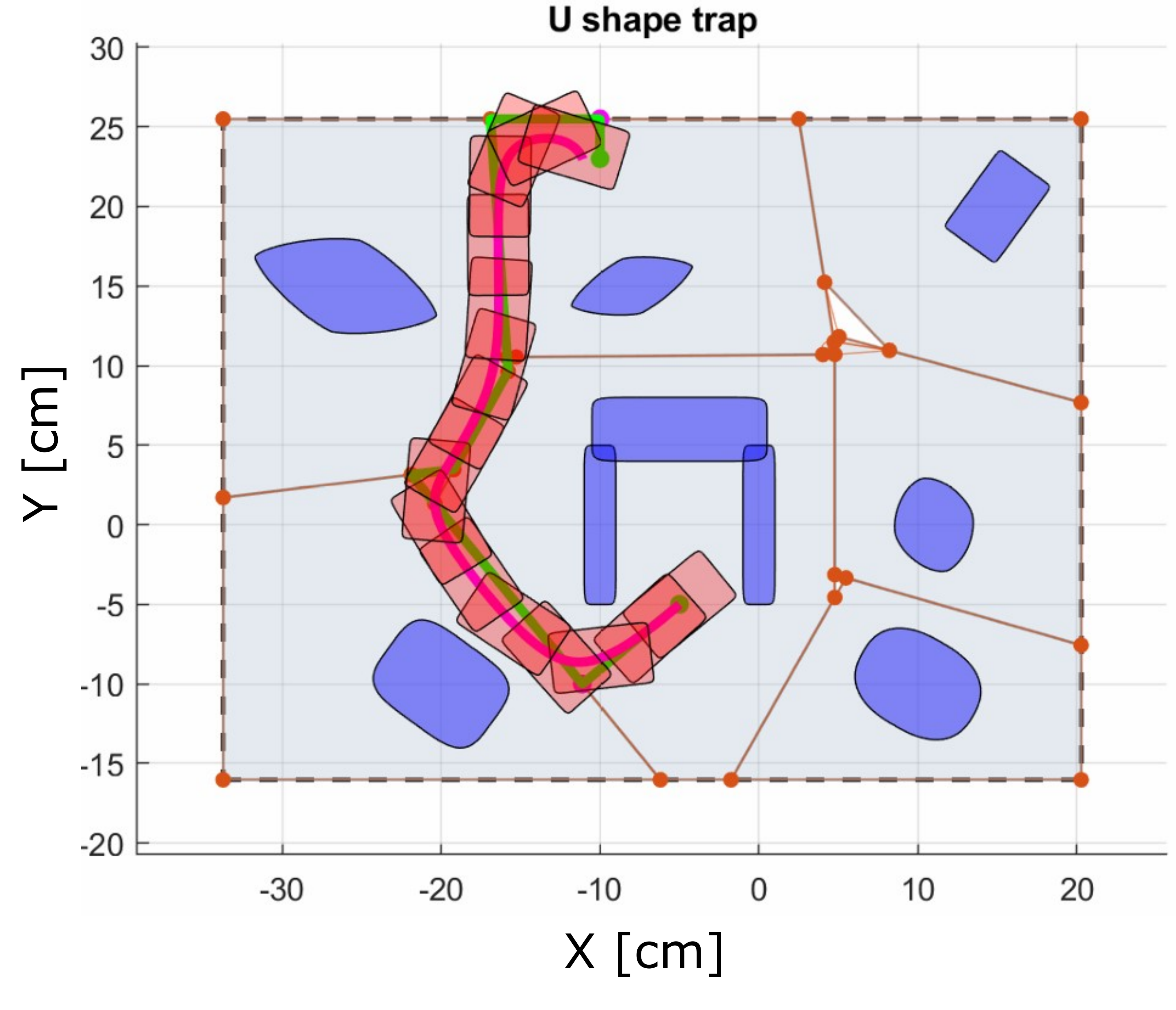}
    \caption[] 
    {\small T block scenario.}
  \end{subfigure}%
\caption{Our method in U-block and T-block scenarios, where closed obstacles are segmented into the same Voronoi region.} 
\label{fig:2DTU}
\end{figure}

To further validate our method, we compare it against three classical motion planners: \textbf{A*}, \textbf{RRT*} and \textbf{Genetic Algorithm} (GA).
We design three 2D benchmark scenarios: \textbf{narrow passage} (Fig. \ref{fig:exp_2D}), \textbf{T-block}, and \textbf{U-block} (Fig. \ref{fig:2DTU}), which are well-known challenging cases for motion planning. The results of our experiments are summarized in Table \ref{tab:2D}, where $[-]$ represents failure in the given time. All computations are performed in MATLAB. We randomly selected start and target points five times in each map to make comparisons.

\begin{table}[h]
    \centering
    \renewcommand{\arraystretch}{1.2} 
    \scriptsize
    \begin{tabular}{|c|c|c|c|c|}
        \hline
        \textbf{Method} & \textbf{Metric} & \textbf{Narrow} & \textbf{T-block} & \textbf{U-block} \\
        \hline
        \multirow{3}{*}{A*} & planning time (s) & 0.42 & 0.16 & 0.21 \\
                               & arc length (m) & 0.41 & 0.42 & 0.38 \\
                               & min-distance (mm) & 0.7 & 1.5 & 1.0 \\
        \hline
        \multirow{3}{*}{RRT*} & planning time (s) & 8.16 & 2.71 & 5.26 \\
                               & arc length (m) & 0.76 & 0.92 & 0.88 \\
                               & min-distance (mm) & 4.1 & 2.1 & 1.18 \\
        \hline
\multirow{3}{*}{GA} & planning time (s) & 5.78 & 2.94 & - \\
                               & arc length (m) & 0.45 & 0.57 & - \\
                               & min-distance (mm) & 1.1 & 2.3 & - \\
        \hline
        \multirow{3}{*}{Proposed} & planning time (s) & 0.065 & 0.095 & 0.083 \\
                               & arc length (m) & 0.61 & 0.52 & 0.51 \\
                               & min-distance (mm) & 14.2 & 23.5 & 26.3 \\                  
        \hline
    \end{tabular}
    \caption{Benchmark for 2D scenario.}
    \label{tab:2D}
\end{table}

In this comparison, we evaluate three key metrics: planning time, arc length, and minimum distance. Planning time refers to the duration of the algorithm’s execution, arc length represents the total trajectory length, and min-distance refers to the closest distance between the robot $SQ_r$ to any obstacle $SQ_o$. 

Our results show that our algorithm consistently completes planning within $0.1 s$ for all 2D scenarios, whereas the classical planners require significantly more time. Additionally, our method achieves the largest min-distance from obstacles among all tested planners. This advantage stems from the Voronoi diagram's inherent property of maximum clearance path (Eq. \ref{eq:vo}), which is particularly crucial in the tasks involving small objects. Given that perception errors are typically in the millimeter range \cite{morar2020comprehensive}, maintaining a safe clearance is essential. In contrast, the other three classical planners frequently generate trajectories very close to obstacles (less than 10 mm), increasing the risk of collisions. Our method, however, consistently maintains a clearance of over 10 mm, ensuring safer navigation.

\section{Simulation: Drone Navigation in 3D Environments} \label{exp3D}

In this section, we apply our algorithm into 3D world. A relevant example is drone navigation in a crowded scenario, where all obstacles, including the drone itself (modeled as an ellipsoid), can be approximated by SQ.

\subsection{Experiment setup}
We designed four challenging scenarios: narrow-passage, trap, complex, and common:
$a)$ Narrow-passage scenario (Fig. \ref{fig:pillar}): This setup consists of several pillar-like obstacles, a common real-world configuration \cite{gao2020teach}. We manually position four SQs as pillars, creating a passage just wide enough for the drone to pass through. Specifically, the gap between obstacles is smaller than the drone’s long axis but larger than its short axis, requiring the drone to adjust its orientation to navigate successfully. The other two passages are too narrow for passage.
$b)$ Trap scenario (Fig. \ref{fig:trap}): The drone starts inside a confined space, or “trap,” with only one exit. The opening is positioned on the side opposite to the target location, making path planning more challenging.
$c)$ Complex scenario (Fig. \ref{fig:complex}): This environment contains a large number of obstacles, significantly increasing the difficulty of finding a feasible trajectory.
$d)$ Common scenario: A simpler environment with fewer obstacles, representing a common case for comparison.

We compare our method against three classical motion planners: A*, RRT*, and GA as in Section \ref{sec:2D}. Additionally, we evaluate our approach against GCOPTER, a cutting-edge algorithm, using its open-source implementation \cite{WANG2022GCOPTER}. To integrate GCOPTER, we first generate the map in MATLAB and convert it into a point cloud, which serves as input for GCOPTER’s environment in Ubuntu 20.04. The results are then collected and analyzed in MATLAB. For each scenario, we randomly select start and target positions five times to ensure a robust comparison.



\subsection{Result analysis}
\begin{figure}[!h]  
\centering
  \begin{subfigure}{0.35\textwidth}
    \includegraphics[width=\linewidth]{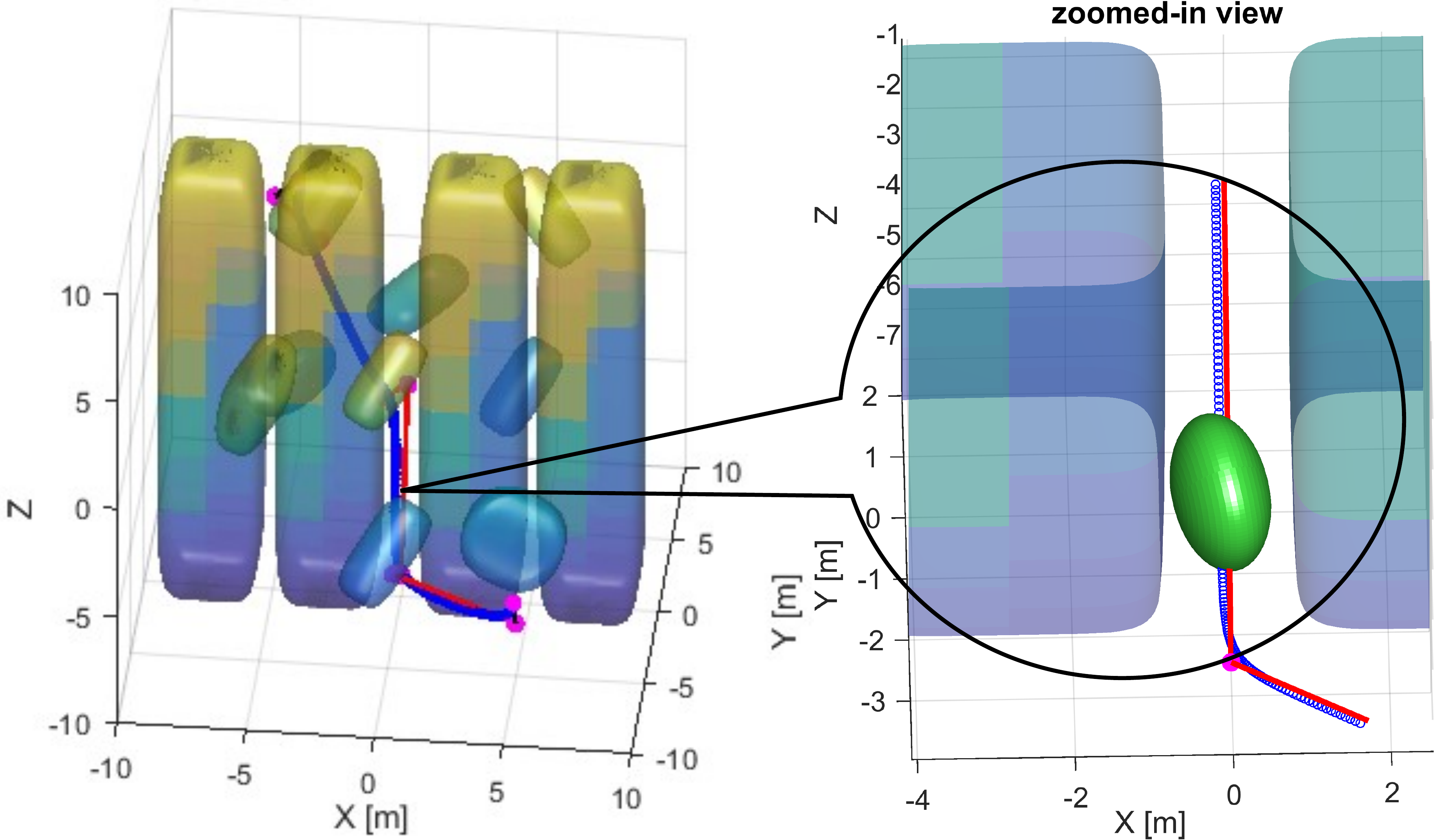}
    \caption[]
    {\small Our method in the narrow-passage scenario, where the drone tilts to navigate through a tight gap between obstacles.}
    \label{fig:pillar}
  \end{subfigure}%
  \vfill  
\begin{subfigure}{0.35\textwidth}
    \includegraphics[width=\linewidth]{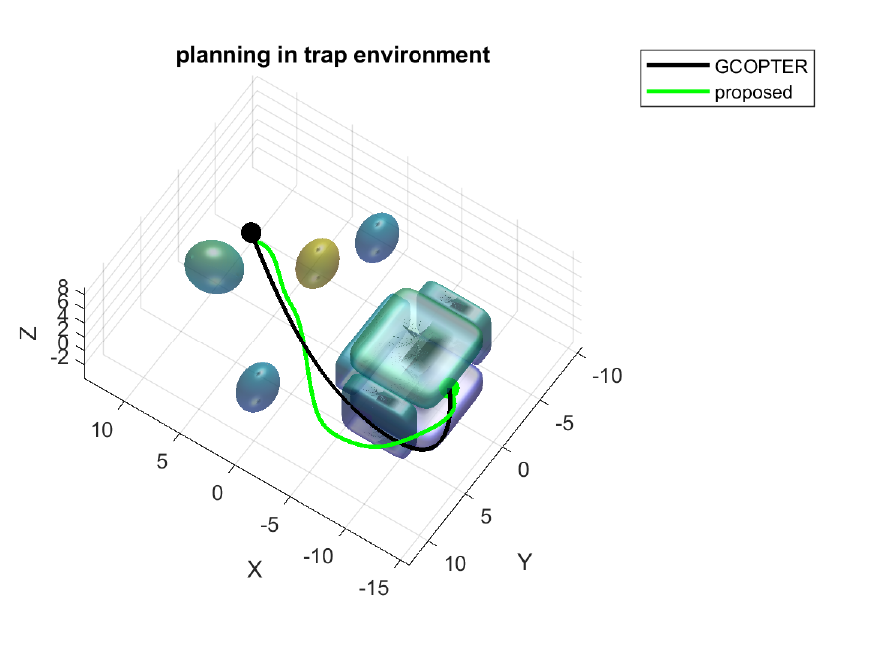}
    \caption[]
    {\small Comparison of our method vs. GCOPTER in the trap scenario, where classical planners tend to fail.}
    \label{fig:trap}
  \end{subfigure}%
  \vfill  
  \begin{subfigure}{0.35\textwidth}
    \includegraphics[width=\linewidth]{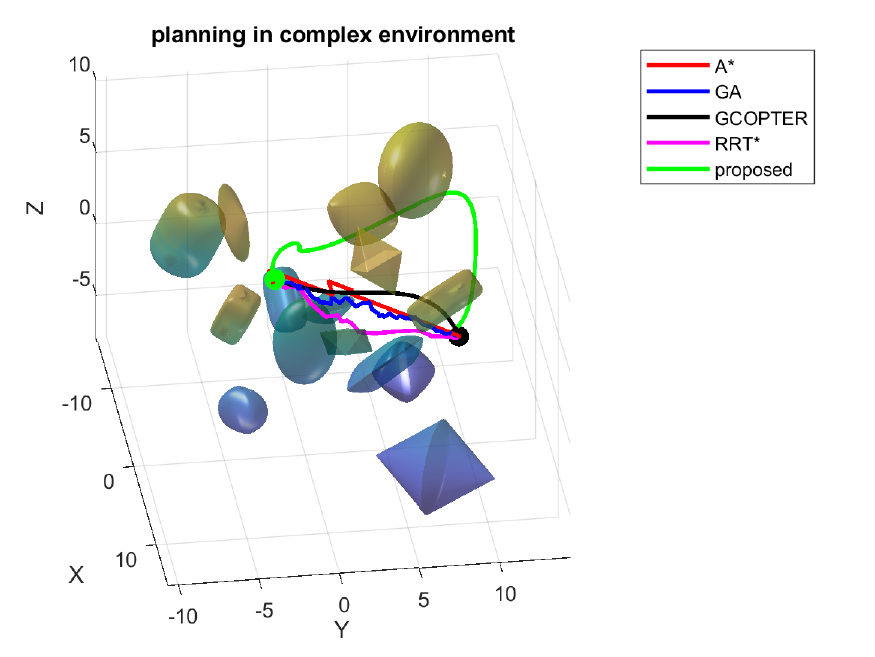}
    \caption[] 
    {\small Comparison of our method vs. GCOPTER and classical planners in the complex scenario.}
    \label{fig:complex}
  \end{subfigure}%
\caption{Comparison of our method with GCOPTER and classical planners in various 3D scenarios.}
\label{fig:cp3D}
\end{figure}

The results are summarized in Table \ref{tab:3D}, where $[-]$ represents failure in the given time. and some representative results are visualized in Fig.\ref{fig:cp3D}. we get the following conclusions: 

\begin{table}[h]
    \centering
    \renewcommand{\arraystretch}{1.1} 
    \scriptsize
    \begin{tabular}{|c|c|c|c|c|c|}
        \hline
        \textbf{Method} & \textbf{Metric} & \textbf{Narrow} & \textbf{Trap} & \textbf{Complex} & \textbf{Common}\\
        \hline
        \multirow{3}{*}{A*} & planning time (s) & 3.58 & 3.77 & 4.25 & 8.06\\
                               & arc length (m) & 22.4 & 28.8 & 29.0 & 42.8\\
                             & min-distance (m) & 0.12 & 0.25 & 0.65 & 0.49\\
        \hline
     \multirow{3}{*}{RRT*} & planning time (s) & 8.92 & - & 1.39 & 1.44\\
                              & arc length (m) & 21.4 & - & 28.8 & 43.5\\
                             & min-distance (m) & 0.22 & - & 0.62 & 0.51\\
        \hline
         \multirow{3}{*}{GA} & planning time (s) & 138.0 & - & 24.89 & 27.5\\
                                & arc length (m) & 24.3 & - & 30.1 & 44.9\\
                               & min-distance (m) & 0.23 & - & 0.71 & 0.55\\
   \hline                            
   \multirow{3}{*}{GCOPTER*} & planning time (s) & - & 0.032 & 0.034 & 0.028\\
                                & arc length (m) & - & 30.2 & 27.4 & 41.3\\
                               & min-distance (m) & - & 0.62 & 0.87 & 0.53\\
        \hline
 \multirow{3}{*}{Proposed} & planning time (s) & 0.26 & 0.28 & 0.42 & 0.16 \\
                               & arc length (m) & 33.47 & 39.3 & 32.0 & 51.2\\
                            & min-distance (m) & 0.28 & 0.95 & 1.46 & 1.71\\                  
        \hline
    \end{tabular}
    \caption{Benchmark for 3D scenario.}
    \label{tab:3D}
\end{table}

\begin{itemize}
    \item Safer path selection: Our method consistently selects paths that keep away from obstacles due to the maximum clearance property (Eq. \ref{eq:vo}). As shown in Table \ref{tab:3D}, our method achieves the largest minimum distance from obstacles compared to other planners. This behavior is particularly evident in the Complex scenario (Fig. \ref{fig:complex}), where our algorithm prioritizes a longer but safer route, avoiding the densely packed obstacles in the center. A similar trend is observed in the Trap scenario (Fig. \ref{fig:trap}), where our method takes a slight detour to safely exit the trap, in contrast to GCOPTER, which follows a more direct yet riskier path.

    \item Comparison with GCOPTER: Our method successfully identifies and eliminates impassable gaps while correctly selecting the only viable narrow passage, as shown in Fig. \ref{fig:pillar}. The drone also tilts at an angle to navigate through the passage, as seen in the zoomed-in view of Fig. \ref{fig:pillar}. In this figure, the green ellipsoid represents the drone, while the transparent shapes depict surrounding obstacles. Our planner ensures that the drone’s short axis aligns with the Voronoi hyperplane normal (Eq. \ref{eq:ori_3D}), enabling it to pass through tight spaces efficiently.
    In contrast, the open-source GCOPTER algorithm does not achieve similar results in the same environment. One limitation of our method is that it is slightly slower than GCOPTER, though both complete execution within less than 0.5 second. This difference may stem from language efficiency, as GCOPTER is implemented in C++, whereas our approach is written in MATLAB.
    
    \item Comparison with classical planners: Our method successfully navigates the Trap scenario by treating trap-like structures as Voronoi regions $V_i$ (Eq.\ref{eq:V_i}), allowing the planner to escape efficiently. In contrast, classical methods such as RRT* and GA fail to exit the trap within the given time limit, as shown in Fig. \ref{fig:trap}. While A* eventually finds an escape route, it requires significantly more time. In general, our method proves much more efficient in 3D environments than classical planners. While RRT*, GA, and A* require more than 1 second, our algorithm completes planning in under 0.5 second, demonstrating its computational efficiency.

\end{itemize}

\textbf{Limitation:} 
Although our approach can handle moderate non-convexity by clustering adjacent obstacles, it faces challenges in highly non-convex environments. Specifically, when the obstacle configurations become too intricate or require merging regions, the current formulation is constrained by the convex polygonal representation in Bensolve. This limitation prevents merging Voronoi regions to form a non-convex space.

\section{Conclusion}

This study introduces an effective path planning approach that integrates SQ modeling with Voronoi diagrams to achieve maximum clearance navigation. By leveraging the differentiable SQ formulation, the computation of Voronoi hyperplanes becomes efficient, thereby enhancing both safety and orientation guidance. Moreover, the flexibility of SQ representations allows the method to approximate a broad range of robot and object geometries (e.g., trap) within a single Voronoi region. Furthermore, by incorporating Minkowski sums, our approach eliminates impractical narrow gaps by merging close obstacles into a same Voronoi region. Due to its mathematical simplicity and generalizability, the proposed method is applicable to both 2D and 3D environments, making it a versatile solution for complex path planning tasks.



We validated our algorithm in 2D and 3D scenarios, including narrow passages, traps, and complex obstacle configurations. Our approach was compared with classical planners (A*, RRT*, GA) and a cutting-edge drone planner (GCOPTER). The results demonstrate that our method is faster than classical methods and effectively navigates through traps and narrow passages, whereas other planners exhibit various limitations. 

In conclusion, the integration of Voronoi diagrams with superquadric representations offers a promising approach for enhancing path planning capabilities in robotics and autonomous systems, particularly in environments that require precise and safe navigation. 
Future work will focus on addressing our limitation by exploring ways to handle non-convex Voronoi regions in highly intricate environments. 
Additionally, we will investigate how to reduce unnecessary turns while keep safety. Meanwhile, we aim to extend the planner to include robotic manipulators, allowing for both end-effector and manipulator path planning to achieve self-collision avoidance.


\bibliographystyle{ieeetr}
\bibliography{ref}


\end{document}